\documentclass[10pt,twocolumn,letterpaper]{article}

\usepackage[pagenumbers]{cvpr} 

\usepackage[dvipsnames]{xcolor}

\definecolor{cvprblue}{rgb}{0.21,0.49,0.74}
\usepackage[pagebackref,breaklinks,colorlinks,citecolor=cvprblue]{hyperref}

\def\paperID{*****} 
\def\confName{CVPR}
\def\confYear{2024}

\title{Efficient Exploration of Image Classifier Failures with Bayesian Optimization and Text-to-Image Models}

\author{Adrien Le Coz$^{1,2}$ \quad Houssem Ouertatani$^{1,3}$ \quad Stéphane Herbin$^2$ \quad Faouzi Adjed$^1$\\
$^1$IRT SystemX, 91120 Palaiseau, France \\ $^2$DTIS, ONERA, Université Paris Saclay F-91123 Palaiseau - France \\ $^3$INRIA Lille - France\\
{\tt\small \{adrien.le-coz, houssem.ouertatani, faouzi.adjed\}@irt-systemx.fr stephane.herbin@onera.fr} 
}

\usepackage{algorithm}
\usepackage{algpseudocode}
\usepackage{graphicx}
\usepackage{subcaption}
\usepackage[utf8]{inputenc}
\usepackage{pgfplots}
\usepgfplotslibrary{groupplots,dateplot}
\usetikzlibrary{external,patterns,shapes.arrows}
\pgfplotsset{compat=newest}
\usepackage{dblfloatfix}    

\begin{document}
\maketitle

\begin{abstract}
    Image classifiers should be used with caution in the real world. Performance evaluated on a validation set may not reflect performance in the real world. In particular, classifiers may perform well for conditions that are frequently encountered during training, but poorly for other infrequent conditions. In this study, we hypothesize that recent advances in text-to-image generative models make them valuable for benchmarking computer vision models such as image classifiers: they can generate images conditioned by textual prompts that cause classifier failures, allowing failure conditions to be described with textual attributes. However, their generation cost becomes an issue when a large number of synthetic images need to be generated, which is the case when many different attribute combinations need to be tested. We propose an image classifier benchmarking method as an iterative process that alternates image generation, classifier evaluation, and attribute selection. This method efficiently explores the attributes that ultimately lead to poor behavior detection.
\end{abstract}

\begin{figure*}[ht]
\centering
\includegraphics[width=\linewidth]{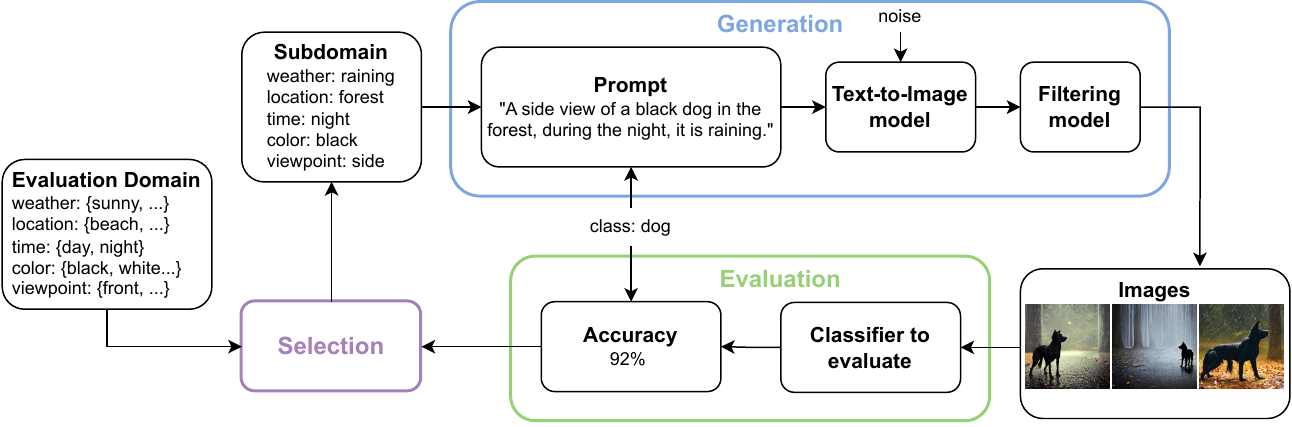}
\caption{Illustration of our method that alternates generation, evaluation, and selection. The selection function selects the next subdomain to evaluate, based on the feedback of the previous subdomains evaluated. With the right choice of selection function, an efficient exploration of the evaluation domain is achieved.
}
\label{fig:diagram}
\end{figure*}

\section{Introduction}
\label{sec:intro}

In computer vision, deep learning models have achieved remarkable successes, consistently pushing the boundaries of what's possible in image classification \cite{he2016deep}, object detection \cite{redmon2016you}, and many other applications. 
Despite these achievements, a persistent challenge remains: accurately discerning when the predictions made by these models can be trusted \cite{amodei2016concrete}. This is especially important for critical decision systems such as autonomous vehicles or medical imaging diagnostics. Even for less critical systems, errors have a cost that can be financial or reputational.
The reliability of model predictions becomes particularly nebulous under conditions of data shift, inherent biases, and the presence of out-of-distribution (OOD) samples. 
Using pre-trained models can worsen those issues because the pre-training process and data might be unknown. It has been shown that deep neural networks often rely on spurious correlations for making predictions \cite{geirhos2020shortcut}.
The conventional metric of a single accuracy number falls significantly short of comprehensively evaluating a model's performance. It is only a global evaluation of a given data distribution. New benchmarking tools are required.\\
Recently, there have been massive improvements in multimodal models, especially those combining textual and visual data like Text-to-Image generative models. These models have demonstrated an exceptional ability to understand and generate content that captures the nuanced interplay between text and images \cite{ramesh2022hierarchical,saharia2022photorealistic}. This allows new ways of benchmarking image classifiers with generative models. Classifier performance can be studied in relation to the textual attributes of the data \cite{wiles2022discovering,metzen2023identification,vendrow2023dataset}. 
Despite their potential, however, the practical utility of these generative models is limited by the computationally intensive inference process of the underlying diffusion models. For example, in \cite{wiles2022discovering}, testing whether the presence of a flower in an image causes the classifier to sometimes mistake flies for bees required hardware with 20 $\times$ 4 TPUs.\\\
\cite{metzen2023identification} developed a classifier evaluation process that starts with an Operational Design Domain \cite{czarnecki2018operational} that textually describes the conditions the model is likely to encounter during use. It consists of many different combinations of attributes. To test a combination, they use synthetic data from a text-to-image model. They then identify which of these combinations lead to classifier errors. However, a major limitation is the combinatorial explosion: they need to limit the number of evaluated combinations. They suggest using combinatorial testing \cite{nie2011survey}, but we found it far from optimal and not much better than random selection. 
In this paper, we are inspired by the principles of Bayesian Optimization (BO), a black-box global optimization method that is particularly well suited for functions with expensive evaluations. Among others, it has been successfully applied to Neural Architecture Search (NAS) \cite{kandasamy2018neural}.
We propose a novel approach to efficiently explore the semantic attributes of data that most significantly impact classification performance. By leveraging the insights gained from the multimodal models and addressing the limitations imposed by the computational demands of diffusion models, our approach seeks to enhance the reliability and interpretability of computer vision models. This paper details our methodology, which combines the strengths of Bayesian Optimization with the latest advancements in benchmarking computer vision with generative models. This offers a more efficient way to understand classification performance in relation to textual descriptions.

Our contributions are:
\begin{itemize}
    \item We improve the efficiency of using Text-to-Image models to identify the textual attributes leading to classifier failure.
    \item We show that this approach significantly reduces the execution time while performing better than baselines.
    \item We demonstrate through some examples how our approach improves the understanding of classifier failures.
\end{itemize}

\section{Related Work}
\label{sec:related_work}

\paragraph{Text-to-image generative models}
Diffusion models, an essential class of generative models, simulate the process of adding noise to data and then learning to reverse this process, enabling the generation of high-quality data samples. Introduced by \cite{sohl2015deep}, these models have paved the way for advancements in generative modeling by demonstrating how data distribution can be captured through denoising steps.
The development of Denoising Diffusion Probabilistic Models (DDPMs) \cite{ho2020denoising} marked a significant leap forward, refining training and sampling methods to produce high-fidelity images. Building upon these foundations, \cite{dhariwal2021diffusion} introduced key improvements in efficiency and sample quality, leading to outperforming previously state-of-the-art generative models like GANs \cite{goodfellow2014generative} and VAEs \cite{kingma2013auto} in image quality and diversity.
Textual conditioning allows for generating complex and diverse images by prompting them with text. Well-known Text-to-Image models include DALL-E 2 \cite{ramesh2022hierarchical} and Imagen \cite{saharia2022photorealistic}. Stable Diffusion \cite{rombach2022high} emphasizes efficiency and scalability. It also makes high-quality text-to-image generation more accessible as it was published in open-source. While GANs can also be conditioned by text \cite{sauer2023stylegan}, the rapid improvements of diffusion models are hard to match. A main limitation of diffusion models is their inference time, requiring many denoising steps to generate an image. This is an important research avenue \cite{song2020denoising,dhariwal2021diffusion}.

\paragraph{Classifier failure discovery}

Discovering failures or bugs in image classification models has recently been studied more and more. 
One can use large labeled datasets and human verification to identify bugs \cite{gao2023adaptive}. To avoid these requirements, other approaches are based on generative models. In particular, leveraging recent Text-to-Image generative models allows linking textual attributes to classification performance. It is possible to identify bugs in a given classifier by generating many images and then clustering and captioning the ones leading to classification failure \cite{wiles2022discovering}. For instance, the presence of a flower in the images augments the chances of misclassification of flies into bees. However, the required computing resources are enormous.
\cite{vendrow2023dataset} personalizes the generation to a specific dataset to create distribution-shifted versions of the dataset. They can be used to test classification models' robustness to shifts. In our work, we can study combinations of shifts leading to failure, or in other words, corner cases.
\cite{metzen2023identification} identifies subgroups of data leading to degraded performance. Starting from an Operational Design Domain defined by domain experts and consisting of several semantic dimensions. An image classifier is tested on selected subgroups of this domain. We take inspiration from this work but derive a guided and efficient exploration of the attributes.


\paragraph{Bayesian optimization}
Bayesian optimization \cite{bo_jones2001taxonomy,bo_garnett_bayesoptbook_2023} is often discussed in the context of Surrogate-Model Based Optimization (SMBO) \cite{bo_smbo_zaefferer2018surrogate}. The aim is to evaluate the costly objective function as few times as possible. To this end, an efficient model is used as its surrogate.
BO typically relies on regression using Gaussian processes (GPs) in a process generally known as Kriging. Despite their ubiquity, thanks to many positive attributes, GPs have certain drawbacks. The most important one is the cubic complexity, making them inefficient as the observed data points increase. Their use is also contingent on selecting a kernel and possibly a distance function. It is however possible to effectively apply the general BO loop with alternative models, such as deep neural networks \cite{bo_snoek2015scalable}, as well as random forests \cite{bo_hutter2011sequential} and Bayesian neural networks \cite{bo_garnett_bayesoptbook_2023}.


\section{Method}
\label{sec:method}

We propose an efficient iterative process to explore the textual attributes leading to classifier failure. We first introduce the background concepts in subsection \ref{subsec:background}; our definitions for the evaluation domain and subdomains in subsection \ref{subsec:domain}; the general pipeline to generate images for the subdomains in subsection \ref{subsec:generate}; and our proposed guided exploration of attributes that matter in subsection \ref{subsec:explore}.

\begin{algorithm}[ht]
\caption{Exploration of image classifier failures}
\label{algo}
\begin{algorithmic}
\State \textbf{Input}:
    \State $D_{to\_eval}$ the evaluation domain
    \State $S_{to\_eval}$ the list of subdomains to evaluate
    \State $f$: the classifier
    \State $g$: the generative model
    \State $h$: the selection function
    \State $n$: the number of allowed evaluations
    \State $S_{eval} = \emptyset$ the dataset of subdomains evaluations
    \State $s_0 \in S_{to\_eval}$: the initial selected subdomain
    
\State \textbf{Explore subdomains}:
\For{$i=0$ \textbf{to} $n$}
    \State \textbf{Generation}
    \State build prompt $p_i$ from selected subdomain $s_i$
    \State $\hat{x_i} \gets g(p_i)$ \Comment generate and filter images from prompt
    \State \textbf{Evaluation}
    \State $\hat{y_i} \gets \arg \max f(\hat{x_i})$ \Comment compute predicted classes
    \State $a_i \gets acc(y, \hat{y})$ \Comment compute classifier accuracy
    \State \textbf{Selection}
    \State $S_{eval} \gets S_{eval} \cup \{(s_i, a_i)\}$ \Comment add result to dataset
    \State $S_{to\_eval} \gets S_{to\_eval} \setminus s_i$ \Comment remove from list
    \State $s_{i+1} = h(S_{eval}, S_{to\_eval})$ \Comment update $h$ and select next
\EndFor
\end{algorithmic}
\end{algorithm}

\begin{table*}[hb]
    \centering
    \begin{tabular}{cccccc|c}
    \toprule
    Subdomain index & Viewpoint & Color & Time & Location & Weather & Classifier accuracy\\
    \midrule
    0&  side&  white&  day&  at the beach&  sunny& 0.98\\
    1&  side&  white&  day&  at the beach&  snowing& 0.94\\
    2&  side&  white&  day&  at the beach&  raining& 0.86\\
    ...&  ...&  ...&  ...&  ...&  ...& ...\\
    1031&  rear&  blue&  night&  in the mountains&  foggy& 0.66\\
    \bottomrule
    \end{tabular}
    \caption{Reference evaluation data. The generation and evaluation steps were pre-computed, and the results were saved in a table. A table look-up replaces these costly steps to compare different selection functions quickly. This also removes the variance in the process.}
    \label{tab:eval_data}
\end{table*}

\subsection{Background}
\label{subsec:background}

\paragraph{Image classifier} To demonstrate our approach without using considerable computing power, we tackle a simplified task: binary classification of images containing dogs. We construct a dog classifier from a classifier pre-trained on ImageNet \cite{imagenet}, a dataset that contains images of animals or everyday objects. Out of the 1000 classes, 119 are different dog breeds. We sum the classifier probabilities of these classes to get the \emph{dog} probability and sum the rest to get the \emph{not-dog} probability.

\paragraph{Text-to-image generative models}

We use diffusion models as a method for generating images from textual descriptions. They are characterized by their ability to produce high-quality images through a process of denoising. The core mechanism involves a forward diffusion process that incrementally adds noise to an image until it becomes indistinguishable from Gaussian noise. The reverse process, iteratively reconstructing the image from noise, is learned during training. 
For Text-to-Image models, the reverse process is conditioned on textual descriptions of the images.
Specifically, on embeddings derived from a pre-trained language model to ensure the generated images align with the provided textual descriptions. This conditioning is usually integrated with cross-attention \cite{attention}.

\subsection{Define the evaluation domain and subdomains}
\label{subsec:domain}

We call \emph{evaluation domain} the ensemble of deployment environment conditions to evaluate. The conditions are described by textual attributes, each containing a finite number of values. They can be categorical or continuous, but we focus on categorical attributes in this work. The domain comprises all the possible attribute value combinations, which we call \emph{subdomains}. The number of subdomains grows exponentially with the number of attributes considered.

As a starting point, we need to define the textual attributes to explore. Expert knowledge is thus required. As we study image classification of natural images of dogs, we define the following attributes and associated values in brackets: weather [sunny, cloudy, raining, snowing], location [at the beach, in the forest, in the city, inside a house, in a garden, in the desert, in the mountains], time [day, night], color [white, black, brown, beige, gray, red, green, blue], and viewpoint [front, side, rear]. Some combinations are not valid and must be removed.

\subsection{Generate data conditioned by attributes}
\label{subsec:generate}

\begin{figure*}[ht]
    \centering
    \begin{subfigure}{.24\textwidth}
        \centering
       \includegraphics[width=\linewidth]{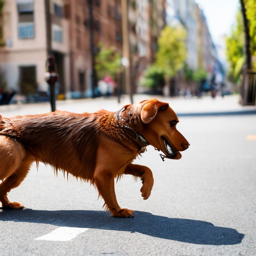}
        \caption{\scriptsize{"A side view of a brown dog in the city, during the day, it is sunny."}}
    \end{subfigure}
    \hfill
    \begin{subfigure}{.24\textwidth}
        \centering
        \includegraphics[width=\linewidth]{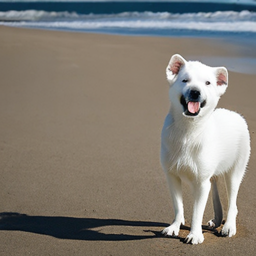}
        \caption{\scriptsize{"A front view of a white dog at the beach, during the day, it is sunny."}}
    \end{subfigure}
    \hfill
    \begin{subfigure}{.24\textwidth}
        \centering
        \includegraphics[width=\linewidth]{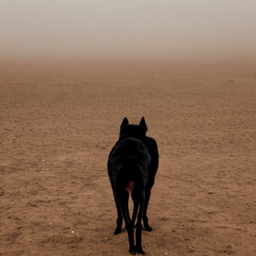}
        \caption{\scriptsize{"A rear view of a black dog in the desert, during the day, it is foggy."}}
    \end{subfigure}
    \hfill
    \begin{subfigure}{.24\textwidth}
        \centering
        \includegraphics[width=\linewidth]{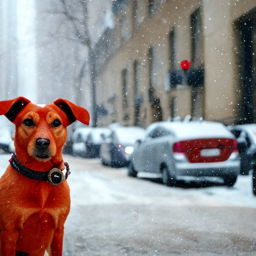}
        \caption{\scriptsize{"A front view of a red dog in the city, during the day, it is snowing."}}
    \end{subfigure}
    
    \begin{subfigure}{.24\textwidth}
        \centering
        \includegraphics[width=\linewidth]{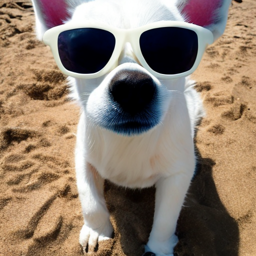}
        \caption{\scriptsize{"A front view of a white dog at the beach, during the day, it is sunny."}}
    \end{subfigure}
    \hfill
    \begin{subfigure}{.24\textwidth}
        \centering
        \includegraphics[width=\linewidth]{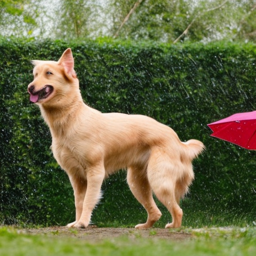}
        \caption{\scriptsize{"A side view of a beige dog in a garden, during the day, it is raining."}}
    \end{subfigure}
    \hfill
    \begin{subfigure}{.24\textwidth}
        \centering
        \includegraphics[width=\linewidth]{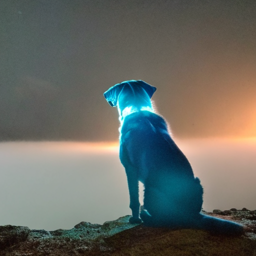}
        \caption{\scriptsize{"A rear view of a blue dog in the mountains, during the night, it is foggy."}}
    \end{subfigure}
    \hfill
    \begin{subfigure}{.24\textwidth}
        \centering
        \includegraphics[width=\linewidth]{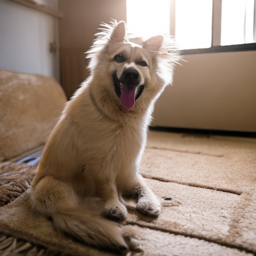}
        \caption{\scriptsize{"A front view of a beige dog inside a house, during the day, it is sunny."}}
    \end{subfigure}

    \caption{Samples of generated images with their associated prompt. Images on the top row are classified as dogs, while those at the bottom are not. Note that some biases of the generative model appear: sunglasses at the beach and an umbrella when raining.}
    \label{fig:samples}
\end{figure*}

\paragraph{Prompt} The first step is to create a textual prompt corresponding to one subdomain attribute. We use a prompt template to fill with the attributes: "A \{viewpoint\} view of a \{color\} dog \{location\}, during the \{time\}, it is \{weather\}.". 

\paragraph{Generate} A Text-to-Image model can then generate images conditioned by the textual prompt. The generation is not deterministic: the starting noisy image is random, and noise is applied to each step of the reverse diffusion process. This means that one textual conditioning leads to a variety of aligned images.

\paragraph{Filter}
The generation is not perfect, and sometimes the synthetic image does not align well with the textual prompt input. We derive a filtering process that follows the generation to limit this issue. We use CLIP \cite{radford2021learning} as a zero-shot subdomain classifier. We have a finite number of subdomains, and each of them is defined as a textual prompt. We thus compute the cosine similarity between a generated image and all subdomains prompts to obtain logits. Applying the softmax function to the logits, we get predicted probabilities that the image corresponds to each subdomain. If the prompt with the maximum probability is indeed the prompt used to generate the image, we consider the image correct otherwise it is filtered out.

\begin{figure*}[h]
\centering
\scalebox{0.75}{
    \large
    \input{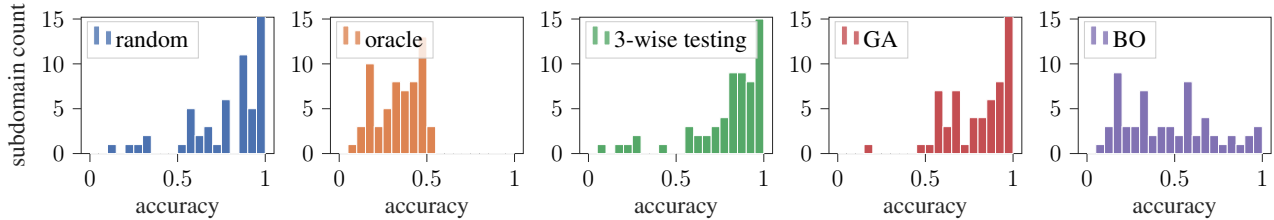}}
\caption{3-wise testing selects 61 subdomains to evaluate. Most of them are high-accuracy. We compare that to the other methods when allowed to explore 61 subdomains. GA and Bayesian optimization identify much more low-accuracy subdomains.}
\label{fig:hist_acc}
\end{figure*}

\subsection{Guided exploration of attributes that matter}
\label{subsec:explore}

Because generating data conditioned by the attributes described above is time-consuming, we propose an efficient exploration of the critical attributes. An iterative process alternates the generation of images for a subdomain, evaluates the classifier on the subdomain, and selects the next subdomain to evaluate based on this feedback. The process is described schematically in Figure \ref{fig:diagram} and more formally in Algorithm \ref{algo}. We propose several selection functions below.

\paragraph{Genetic algorithm (GA)} This is a performant optimization method based on natural selection \cite{holland1992adaptation}. A population of solutions is evaluated. The top performers are preserved, and a crossover operation generates \textit{children} solutions from pairs of \textit{parents}. This new generation of solutions undergoes mutations with a small probability, adding diversity.

\paragraph{Bayesian optimization (BO)}
Our method to efficiently explore the space of subdomains involves the same core loop at the center of Bayesian optimization, relying on a predictive model to guide the search towards the critical subdomains.
\begin{enumerate}
    \item Selection: choose the next subdomain to evaluate using the model
    \item Observation: evaluate the subdomain
    \item Model update: add the new observation to the dataset
\end{enumerate}


The selection policy generally means selecting the point which maximizes an acquisition function. Many acquisition functions exist in the literature, and each presents a different trade-off between exploration and exploitation. The selection policy we use is inspired by Expected Improvement \cite{bo_Mockus1978}, a widely used and generally effective acquisition function. Using the model's estimation of each subdomain's quality, we select the subdomain with the highest potential improvement over the current best subdomain.

\section{Experiments}
\label{sec:exp}

We conduct experiments evaluating the different aspects of our approach. We first provide information on our experimental setting in subsection \ref{subsec:prerequisites_exp}; we provide details on the reference data generation in subsection \ref{subsec:eval_all}; we compare different selection functions, including some baselines, in subsection \ref{subsec:benchmark_selection}; and we display qualitative results of classifier evaluation in \ref{subsec:quali}.

\begin{figure*}[h]
\centering
\begin{subfigure}[t]{0.48\linewidth}
\scalebox{1}{
    \footnotesize
    \input{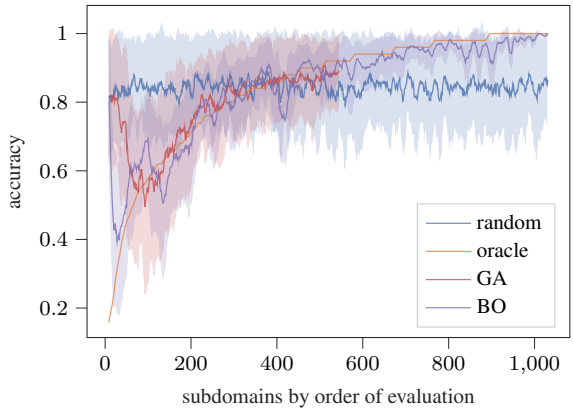}}
\caption{Evolution of the accuracy of selected subdomains during the exploration (lower is better). We used a moving average with a window size of 10 to improve clarity. GA and the BO quickly select low-accuracy subdomains until only higher-accuracy subdomains remain.}
\label{fig:acc}
\end{subfigure}
\hfill
\begin{subfigure}[t]{0.48\linewidth}
\scalebox{1}{
    \footnotesize
    \input{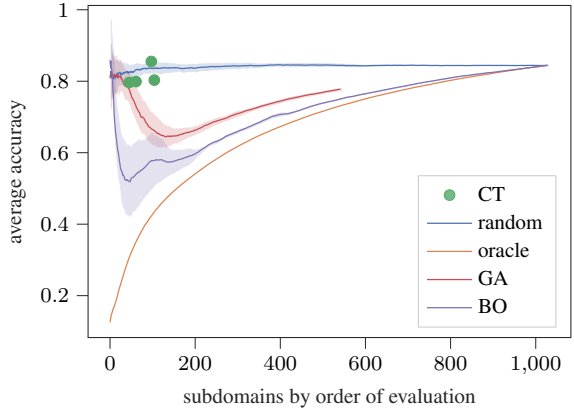}}
\caption{Evolution of the average accuracy on subdomains already evaluated during the exploration (lower is better). All methods converge to the global accuracy. Combinatorial testing is not much better than random selection, compared to the GA and BO.}
\label{fig:avg_acc}
\end{subfigure}
\vfill
\begin{subfigure}[t]{0.48\linewidth}
\scalebox{1}{
    \footnotesize
    \input{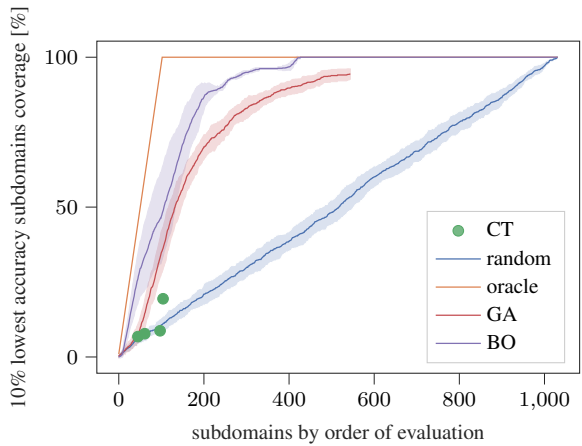}
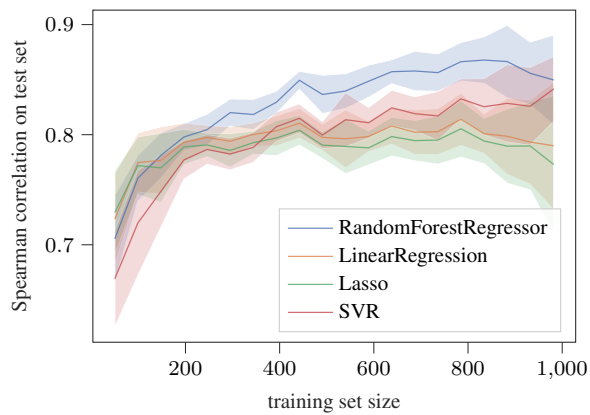}
\caption{Evolution of the 10\% lowest accuracy subdomains coverage (higher is better). We identified the 10\% (103) subdomains with the lowest accuracies and computed what proportion of them is covered by the subdomains selected during the exploration. The BO finds all of them after evaluating $\approx$ 300.}
\label{fig:cov}
\end{subfigure}
\hfill
\begin{subfigure}[t]{0.48\linewidth}
\scalebox{1}{
    \footnotesize
\begin{tikzpicture}

\definecolor{darkslategray38}{RGB}{38,38,38}
\definecolor{indianred1967882}{RGB}{196,78,82}
\definecolor{lightgray204}{RGB}{204,204,204}
\definecolor{mediumseagreen85168104}{RGB}{85,168,104}
\definecolor{peru22113282}{RGB}{221,132,82}
\definecolor{steelblue76114176}{RGB}{76,114,176}

\begin{axis}[
axis line style={darkslategray38},
height=6cm,
legend cell align={left},
legend style={
  fill opacity=0.8,
  draw opacity=1,
  text opacity=1,
  at={(0.97,0.03)},
  anchor=south east,
  draw=lightgray204
},
tick align=outside,
tick pos=left,
width=8cm,
x grid style={lightgray204},
xlabel=\textcolor{darkslategray38}{training set size},
xmin=3.4, xmax=1028.6,
xtick style={color=darkslategray38},
y grid style={lightgray204},
ylabel=\textcolor{darkslategray38}{Spearman correlation on test set},
ymin=0.611474361098132, ymax=0.913148221743342,
ytick style={color=darkslategray38}
]
\path [draw=white, fill=steelblue76114176, opacity=0.2, line width=0.32pt]
(axis cs:50,0.745153915569005)
--(axis cs:50,0.665681256135312)
--(axis cs:99,0.739629809462239)
--(axis cs:148,0.760534650351857)
--(axis cs:197,0.785664536802107)
--(axis cs:246,0.790507007494557)
--(axis cs:295,0.807596089877999)
--(axis cs:344,0.804516627646776)
--(axis cs:393,0.819255824231499)
--(axis cs:442,0.841121065473301)
--(axis cs:491,0.81896215844089)
--(axis cs:540,0.823861508401922)
--(axis cs:589,0.833915335036502)
--(axis cs:638,0.84628358668975)
--(axis cs:687,0.839875597650564)
--(axis cs:736,0.839340164526864)
--(axis cs:785,0.848744255092207)
--(axis cs:834,0.846637089358693)
--(axis cs:883,0.833504124175275)
--(axis cs:932,0.827315838756921)
--(axis cs:982,0.808749173423535)
--(axis cs:982,0.890661540574484)
--(axis cs:982,0.890661540574484)
--(axis cs:932,0.88423482963727)
--(axis cs:883,0.899435773532197)
--(axis cs:834,0.888950584309337)
--(axis cs:785,0.883836590769677)
--(axis cs:736,0.873444784465895)
--(axis cs:687,0.875879720264751)
--(axis cs:638,0.868129147538423)
--(axis cs:589,0.863224437817929)
--(axis cs:540,0.855487849817599)
--(axis cs:491,0.854080549142996)
--(axis cs:442,0.857757764745445)
--(axis cs:393,0.839569669790616)
--(axis cs:344,0.832126720667912)
--(axis cs:295,0.832491825368127)
--(axis cs:246,0.818710479796891)
--(axis cs:197,0.810227634613331)
--(axis cs:148,0.801946945050516)
--(axis cs:99,0.78137698993515)
--(axis cs:50,0.745153915569005)
--cycle;

\path [draw=white, fill=peru22113282, opacity=0.2, line width=0.32pt]
(axis cs:50,0.764502004811161)
--(axis cs:50,0.682082041476465)
--(axis cs:99,0.747709216551352)
--(axis cs:148,0.746161260385457)
--(axis cs:197,0.775346922595502)
--(axis cs:246,0.786467606597656)
--(axis cs:295,0.780945226220507)
--(axis cs:344,0.789656453043912)
--(axis cs:393,0.789744327012943)
--(axis cs:442,0.797014683839521)
--(axis cs:491,0.785591873867306)
--(axis cs:540,0.769831016705109)
--(axis cs:589,0.784561708163092)
--(axis cs:638,0.791386460321404)
--(axis cs:687,0.782090582141732)
--(axis cs:736,0.781868371977298)
--(axis cs:785,0.790179742259201)
--(axis cs:834,0.782972198196578)
--(axis cs:883,0.763984897681344)
--(axis cs:932,0.754957874090505)
--(axis cs:982,0.731804198132683)
--(axis cs:982,0.848037743045267)
--(axis cs:982,0.848037743045267)
--(axis cs:932,0.831673508540913)
--(axis cs:883,0.833158285173215)
--(axis cs:834,0.818812491885378)
--(axis cs:785,0.837854746736374)
--(axis cs:736,0.82362839459298)
--(axis cs:687,0.822441961603986)
--(axis cs:638,0.824557878544654)
--(axis cs:589,0.812159164139087)
--(axis cs:540,0.822720648762926)
--(axis cs:491,0.809549797277546)
--(axis cs:442,0.823944693342921)
--(axis cs:393,0.817120065700311)
--(axis cs:344,0.810874696722663)
--(axis cs:295,0.807163005606646)
--(axis cs:246,0.808474774073383)
--(axis cs:197,0.810681412361566)
--(axis cs:148,0.807199638146653)
--(axis cs:99,0.801645854062993)
--(axis cs:50,0.764502004811161)
--cycle;

\path [draw=white, fill=mediumseagreen85168104, opacity=0.2, line width=0.32pt]
(axis cs:50,0.767436657270246)
--(axis cs:50,0.691360308280168)
--(axis cs:99,0.745697477128357)
--(axis cs:148,0.738415012718086)
--(axis cs:197,0.77272188199771)
--(axis cs:246,0.780015854448531)
--(axis cs:295,0.770497620902155)
--(axis cs:344,0.782211782196155)
--(axis cs:393,0.780964187793053)
--(axis cs:442,0.790520615271578)
--(axis cs:491,0.778305448892863)
--(axis cs:540,0.764519262269052)
--(axis cs:589,0.772579252137039)
--(axis cs:638,0.78056819189538)
--(axis cs:687,0.776250023634838)
--(axis cs:736,0.773122630847846)
--(axis cs:785,0.779576115954106)
--(axis cs:834,0.773603217156665)
--(axis cs:883,0.755909302576526)
--(axis cs:932,0.749773178363108)
--(axis cs:982,0.710300607326938)
--(axis cs:982,0.835521514773813)
--(axis cs:982,0.835521514773813)
--(axis cs:932,0.829754733561232)
--(axis cs:883,0.823333121970779)
--(axis cs:834,0.815007195206901)
--(axis cs:785,0.831102184563884)
--(axis cs:736,0.81718325643246)
--(axis cs:687,0.812957330135806)
--(axis cs:638,0.816028819987783)
--(axis cs:589,0.803501699693161)
--(axis cs:540,0.814209664058781)
--(axis cs:491,0.802689106778526)
--(axis cs:442,0.817576365159159)
--(axis cs:393,0.812888748468131)
--(axis cs:344,0.803329910192673)
--(axis cs:295,0.800950169712833)
--(axis cs:246,0.801330159134968)
--(axis cs:197,0.805098950301079)
--(axis cs:148,0.801289906304215)
--(axis cs:99,0.798053707107699)
--(axis cs:50,0.767436657270246)
--cycle;

\path [draw=white, fill=indianred1967882, opacity=0.2, line width=0.32pt]
(axis cs:50,0.712985284154361)
--(axis cs:50,0.625186809309278)
--(axis cs:99,0.672083561387786)
--(axis cs:148,0.715891519143793)
--(axis cs:197,0.759629543550858)
--(axis cs:246,0.773111794761059)
--(axis cs:295,0.767860505813348)
--(axis cs:344,0.774908105717333)
--(axis cs:393,0.793878065256474)
--(axis cs:442,0.801681983326031)
--(axis cs:491,0.787336794490036)
--(axis cs:540,0.788985957749891)
--(axis cs:589,0.796918100018439)
--(axis cs:638,0.808269397279167)
--(axis cs:687,0.803699364866103)
--(axis cs:736,0.794773687504844)
--(axis cs:785,0.814221109283656)
--(axis cs:834,0.799850306709549)
--(axis cs:883,0.793307926427053)
--(axis cs:932,0.790416275234107)
--(axis cs:982,0.8122007807052)
--(axis cs:982,0.871075755593982)
--(axis cs:982,0.871075755593982)
--(axis cs:932,0.861192451182328)
--(axis cs:883,0.863651982234754)
--(axis cs:834,0.850948784635845)
--(axis cs:785,0.850842873744868)
--(axis cs:736,0.839374191749923)
--(axis cs:687,0.834433181532749)
--(axis cs:638,0.840479441208106)
--(axis cs:589,0.824806034344562)
--(axis cs:540,0.838241101883359)
--(axis cs:491,0.811972473184351)
--(axis cs:442,0.828081050402653)
--(axis cs:393,0.82054138420529)
--(axis cs:344,0.802069041365604)
--(axis cs:295,0.79700287917059)
--(axis cs:246,0.800030638336893)
--(axis cs:197,0.794697190488535)
--(axis cs:148,0.78126198545217)
--(axis cs:99,0.767589649227995)
--(axis cs:50,0.712985284154361)
--cycle;

\addplot [line width=0.48pt, steelblue76114176, opacity=0.8]
table {%
50 0.705417585852159
99 0.760503399698695
148 0.781240797701187
197 0.797946085707719
246 0.804608743645724
295 0.820043957623063
344 0.818321674157344
393 0.829412747011057
442 0.849439415109373
491 0.836521353791943
540 0.83967467910976
589 0.848569886427216
638 0.857206367114086
687 0.857877658957658
736 0.85639247449638
785 0.866290422930942
834 0.867793836834015
883 0.866469948853736
932 0.855775334197096
982 0.849705356999009
};
\addlegendentry{RandomForestRegressor}
\addplot [line width=0.48pt, peru22113282, opacity=0.8]
table {%
50 0.723292023143813
99 0.774677535307172
148 0.776680449266055
197 0.793014167478534
246 0.797471190335519
295 0.794054115913577
344 0.800265574883287
393 0.803432196356627
442 0.810479688591221
491 0.797570835572426
540 0.796275832734018
589 0.79836043615109
638 0.807972169433029
687 0.802266271872859
736 0.802748383285139
785 0.814017244497787
834 0.800892345040978
883 0.798571591427279
932 0.793315691315709
982 0.789920970588975
};
\addlegendentry{LinearRegression}
\addplot [line width=0.48pt, mediumseagreen85168104, opacity=0.8]
table {%
50 0.729398482775207
99 0.771875592118028
148 0.769852459511151
197 0.788910416149394
246 0.790673006791749
295 0.785723895307494
344 0.792770846194414
393 0.796926468130592
442 0.804048490215369
491 0.790497277835694
540 0.789364463163916
589 0.7880404759151
638 0.798298505941581
687 0.794603676885322
736 0.795152943640153
785 0.805339150258995
834 0.794305206181783
883 0.789621212273652
932 0.78976395596217
982 0.772911061050376
};
\addlegendentry{Lasso}
\addplot [line width=0.48pt, indianred1967882, opacity=0.8]
table {%
50 0.66908604673182
99 0.71983660530789
148 0.748576752297982
197 0.777163367019696
246 0.786571216548976
295 0.782431692491969
344 0.788488573541469
393 0.807209724730882
442 0.814881516864342
491 0.799654633837193
540 0.813613529816625
589 0.810862067181501
638 0.824374419243637
687 0.819066273199426
736 0.817073939627384
785 0.832531991514262
834 0.825399545672697
883 0.828479954330903
932 0.825804363208217
982 0.841638268149591
};
\addlegendentry{SVR}
\end{axis}

\end{tikzpicture}
    }
\caption{Spearman's rank correlation coefficient for different predictors and training set sizes. It measures the strength and direction of the monotonic relationship between two ranked variables, here the predicted and test accuracies. A value close to 1 means the relationship between the two variables is monotonic. Except for SVR, all predictors perform similarly well. Lasso is the best method for small training sizes.}
\label{fig:spearman}
\end{subfigure}
\caption{Different metrics to compare the quality of the subdomain selection when iterating on the loop generation, evaluation, and selection. In general, combinatorial testing is not much better than random selection, and it only gives a few options for the number of subdomains selected. GA and BO are much more efficient and can explore any given number of subdomains according to the computation time available. Note that the x-axis of \ref{fig:acc}, \ref{fig:avg_acc}, and \ref{fig:cov} could be replaced by GPU.hours going from 0 to $\approx$ 200 as mentioned in Subsection \ref{subsec:eval_all}. All plots are averages over 10 seeds and the standard deviations are shown.}
\label{fig:acc_plots}
\end{figure*}

\subsection{Prerequisites}
\label{subsec:prerequisites_exp}

\paragraph{Classifier} We study a classifier with the ViT-B/16 \cite{dosovitskiy2020image} architecture. Weights are from torchvision, following a pre-training on the ImageNet dataset. The binary classifier's accuracy on ImageNet validation data is more than 99\%. We want to assess its performance on data that is more diverse than in the original dataset to see if it can generalize well.

\paragraph{Subdomains}
The number of possible attribute combinations is 1 class (dog) $\times$ 4 weathers $\times$ 7 locations $\times$ 2 time periods $\times$ 8 colors $\times$ 3 viewpoints = 1344. However, some of the combinations are impossible (e.g., "during the night, it is sunny" or "in a house, it is snowing"). After filtering those, 1032 combinations remain, forming all the possible subdomains to evaluate.

\paragraph{Generative model} We use Stability AI's implementation of Stable Diffusion 2.1 as a text-to-image generative model.  Its architecture is based on Latent Diffusion Models \cite{rombach2022high}, and text conditioning uses a fixed pre-trained text encoder based on CLIP ViT/H. Generated images have a $512 \times 512$ resolution, but we resize them into $256 \times 256$ to save disk space. Resizing images at a lower resolution is part of the classifier data preprocessing anyway. We treat this model as a black box transforming textual input prompts into diverse corresponding images.

\paragraph{Filtering model} Because the generation is imperfect, we need to filter out generated images that do not align well with the textual input prompt. We use a subdomain classifier that classifies generated images into one of the subdomains. This classifier is a pre-trained CLIP ViT-L/14 adapted as a zero-shot classifier.

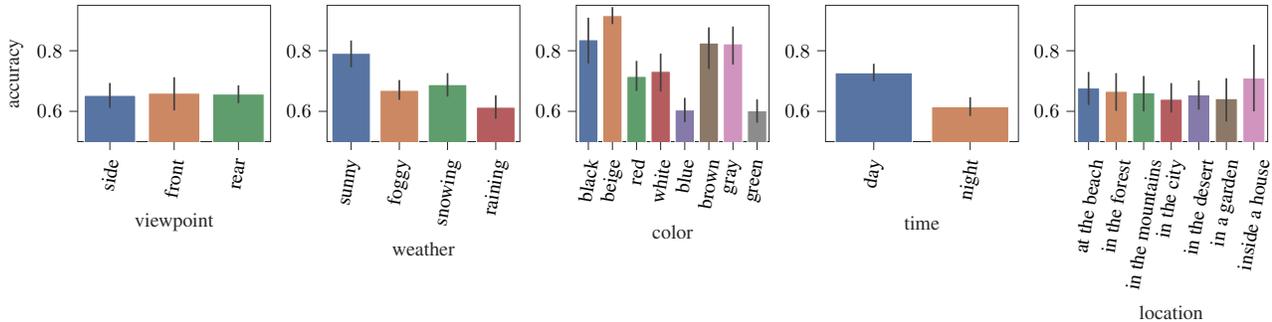
\begin{figure*}[ht]
    \centering
    \scalebox{0.75}{
\begin{tikzpicture}

\definecolor{darkslategray38}{RGB}{38,38,38}
\definecolor{darkslategray66}{RGB}{66,66,66}
\definecolor{gray140}{RGB}{140,140,140}
\definecolor{gray140120102}{RGB}{140,120,102}
\definecolor{indianred1819295}{RGB}{181,92,95}
\definecolor{lightgray204}{RGB}{204,204,204}
\definecolor{lightslategray133122170}{RGB}{133,122,170}
\definecolor{mediumseagreen95157109}{RGB}{95,157,109}
\definecolor{peru20313699}{RGB}{203,136,99}
\definecolor{plum208148190}{RGB}{208,148,190}
\definecolor{steelblue88116163}{RGB}{88,116,163}

\begin{groupplot}[group style={group size=5 by 1}]
\nextgroupplot[
axis line style={darkslategray38},
height=4cm,
tick align=outside,
tick pos=left,
width=5cm,
x grid style={lightgray204},
xlabel=\textcolor{darkslategray38}{viewpoint},
xmin=-0.5, xmax=2.5,
xtick style={color=darkslategray38},
xtick={0,1,2},
xticklabel style={rotate=80.0},
xticklabels={side,front,rear},
y grid style={lightgray204},
ylabel=\textcolor{darkslategray38}{accuracy},
ymin=0.5, ymax=0.95,
ytick style={color=darkslategray38}
]
\draw[draw=white,fill=steelblue88116163,line width=0.32pt] (axis cs:-0.4,0) rectangle (axis cs:0.4,0.652599982917309);
\draw[draw=white,fill=peru20313699,line width=0.32pt] (axis cs:0.6,0) rectangle (axis cs:1.4,0.659999983112017);
\draw[draw=white,fill=mediumseagreen95157109,line width=0.32pt] (axis cs:1.6,0) rectangle (axis cs:2.4,0.656959983944893);
\addplot [line width=0.864pt, darkslategray66]
table {%
0 0.611779980607331
0 0.693614981163293
};
\addplot [line width=0.864pt, darkslategray66]
table {%
1 0.603193317602078
1 0.712553315420945
};
\addplot [line width=0.864pt, darkslategray66]
table {%
2 0.627199985259771
2 0.68576798183918
};

\nextgroupplot[
axis line style={darkslategray38},
height=4cm,
tick align=outside,
tick pos=left,
width=5cm,
x grid style={lightgray204},
xlabel=\textcolor{darkslategray38}{weather},
xmin=-0.5, xmax=3.5,
xtick style={color=darkslategray38},
xtick={0,1,2,3},
xticklabel style={rotate=80.0},
xticklabels={sunny,foggy,snowing,raining},
y grid style={lightgray204},
ymin=0.5, ymax=0.95,
ytick style={color=darkslategray38}
]
\draw[draw=white,fill=steelblue88116163,line width=0.32pt] (axis cs:-0.4,0) rectangle (axis cs:0.4,0.791538444849161);
\draw[draw=white,fill=peru20313699,line width=0.32pt] (axis cs:0.6,0) rectangle (axis cs:1.4,0.669275346873463);
\draw[draw=white,fill=mediumseagreen95157109,line width=0.32pt] (axis cs:1.6,0) rectangle (axis cs:2.4,0.688275846941718);
\draw[draw=white,fill=indianred1819295,line width=0.32pt] (axis cs:2.6,0) rectangle (axis cs:3.4,0.613605424457667);
\addplot [line width=0.864pt, darkslategray66]
table {%
0 0.745346135359544
0 0.833846133546187
};
\addplot [line width=0.864pt, darkslategray66]
table {%
1 0.637963752783295
1 0.703188390446746
};
\addplot [line width=0.864pt, darkslategray66]
table {%
2 0.64861205514392
2 0.726206880191277
};
\addplot [line width=0.864pt, darkslategray66]
table {%
3 0.575625835333754
3 0.652792498063879
};

\nextgroupplot[
axis line style={darkslategray38},
height=4cm,
tick align=outside,
tick pos=left,
width=5cm,
x grid style={lightgray204},
xlabel=\textcolor{darkslategray38}{color},
xmin=-0.5, xmax=7.5,
xtick style={color=darkslategray38},
xtick={0,1,2,3,4,5,6,7},
xticklabel style={rotate=80.0},
xticklabels={black,beige,red,white,blue,brown,gray,green},
y grid style={lightgray204},
ymin=0.5, ymax=0.95,
ytick style={color=darkslategray38}
]
\draw[draw=white,fill=steelblue88116163,line width=0.32pt] (axis cs:-0.4,0) rectangle (axis cs:0.4,0.836363619024103);
\draw[draw=white,fill=peru20313699,line width=0.32pt] (axis cs:0.6,0) rectangle (axis cs:1.4,0.915999978780746);
\draw[draw=white,fill=mediumseagreen95157109,line width=0.32pt] (axis cs:1.6,0) rectangle (axis cs:2.4,0.715319130014866);
\draw[draw=white,fill=indianred1819295,line width=0.32pt] (axis cs:2.6,0) rectangle (axis cs:3.4,0.731999981403351);
\draw[draw=white,fill=lightslategray133122170,line width=0.32pt] (axis cs:3.6,0) rectangle (axis cs:4.4,0.604523796233393);
\draw[draw=white,fill=gray140120102,line width=0.32pt] (axis cs:4.6,0) rectangle (axis cs:5.4,0.825714247567313);
\draw[draw=white,fill=plum208148190,line width=0.32pt] (axis cs:5.6,0) rectangle (axis cs:6.4,0.822857115949903);
\draw[draw=white,fill=gray140,line width=0.32pt] (axis cs:6.6,0) rectangle (axis cs:7.4,0.601680656816779);
\addplot [line width=0.864pt, darkslategray66]
table {%
0 0.75818179954182
0 0.909090898253701
};
\addplot [line width=0.864pt, darkslategray66]
table {%
1 0.88799996972084
1 0.943999981880188
};
\addplot [line width=0.864pt, darkslategray66]
table {%
2 0.667212746672808
2 0.766414874047041
};
\addplot [line width=0.864pt, darkslategray66]
table {%
3 0.665333318610986
3 0.790666647851467
};
\addplot [line width=0.864pt, darkslategray66]
table {%
4 0.564505940325381
4 0.644529745945086
};
\addplot [line width=0.864pt, darkslategray66]
table {%
5 0.739999966961997
5 0.877142821039472
};
\addplot [line width=0.864pt, darkslategray66]
table {%
6 0.754214247848306
6 0.87999997820173
};
\addplot [line width=0.864pt, darkslategray66]
table {%
7 0.562516792387772
7 0.639327716072943
};

\nextgroupplot[
axis line style={darkslategray38},
height=4cm,
tick align=outside,
tick pos=left,
width=5cm,
x grid style={lightgray204},
xlabel=\textcolor{darkslategray38}{time},
xmin=-0.5, xmax=1.5,
xtick style={color=darkslategray38},
xtick={0,1},
xticklabel style={rotate=80.0},
xticklabels={day,night},
y grid style={lightgray204},
ymin=0.5, ymax=0.95,
ytick style={color=darkslategray38}
]
\draw[draw=white,fill=steelblue88116163,line width=0.32pt] (axis cs:-0.4,0) rectangle (axis cs:0.4,0.727272709933194);
\draw[draw=white,fill=peru20313699,line width=0.32pt] (axis cs:0.6,0) rectangle (axis cs:1.4,0.61515787855575);
\addplot [line width=0.864pt, darkslategray66]
table {%
0 0.699272710796107
0 0.756913619122722
};
\addplot [line width=0.864pt, darkslategray66]
table {%
1 0.584723668190602
1 0.646213142107192
};

\nextgroupplot[
axis line style={darkslategray38},
height=4cm,
tick align=outside,
tick pos=left,
width=5cm,
x grid style={lightgray204},
xlabel=\textcolor{darkslategray38}{location},
xmin=-0.5, xmax=6.5,
xtick style={color=darkslategray38},
xtick={0,1,2,3,4,5,6},
xticklabel style={rotate=80.0},
xticklabels={
  at the beach,
  in the forest,
  in the mountains,
  in the city,
  in the desert,
  in a garden,
  inside a house
},
y grid style={lightgray204},
ymin=0.5, ymax=0.95,
ytick style={color=darkslategray38}
]
\draw[draw=white,fill=steelblue88116163,line width=0.32pt] (axis cs:-0.4,0) rectangle (axis cs:0.4,0.67699998691678);
\draw[draw=white,fill=peru20313699,line width=0.32pt] (axis cs:0.6,0) rectangle (axis cs:1.4,0.666122433178279);
\draw[draw=white,fill=mediumseagreen95157109,line width=0.32pt] (axis cs:1.6,0) rectangle (axis cs:2.4,0.661481462419033);
\draw[draw=white,fill=indianred1819295,line width=0.32pt] (axis cs:2.6,0) rectangle (axis cs:3.4,0.639999982485404);
\draw[draw=white,fill=lightslategray133122170,line width=0.32pt] (axis cs:3.6,0) rectangle (axis cs:4.4,0.654509789803449);
\draw[draw=white,fill=gray140120102,line width=0.32pt] (axis cs:4.6,0) rectangle (axis cs:5.4,0.642051262733264);
\draw[draw=white,fill=plum208148190,line width=0.32pt] (axis cs:5.6,0) rectangle (axis cs:6.4,0.709999978542328);
\addplot [line width=0.864pt, darkslategray66]
table {%
0 0.620999991856515
0 0.730012487675995
};
\addplot [line width=0.864pt, darkslategray66]
table {%
1 0.601622434135298
1 0.726122433213251
};
\addplot [line width=0.864pt, darkslategray66]
table {%
2 0.599916646188056
2 0.716675906807736
};
\addplot [line width=0.864pt, darkslategray66]
table {%
3 0.595992290933545
3 0.693230750939021
};
\addplot [line width=0.864pt, darkslategray66]
table {%
4 0.605490182500844
4 0.701990179396143
};
\addplot [line width=0.864pt, darkslategray66]
table {%
5 0.566666653876503
5 0.709256389899514
};
\addplot [line width=0.864pt, darkslategray66]
table {%
6 0.599999964237213
6 0.819999992847443
};
\end{groupplot}

\end{tikzpicture}}
    \caption{Average accuracies for each value of each attribute. The 95\% confidence interval is also shown.}
    \label{fig:barplot}
\end{figure*}

\paragraph{Baselines}
For comparison, we include some methods of selecting the subdomains to evaluate as baselines.\\
\begin{itemize}
    \item The \emph{random selection} simply randomly picks a subdomain to test in the list of the remaining ones.
    \item The \emph{oracle} knows all the subdomain's accuracies in advance, and it chooses the subdomains by order of increasing accuracy. This is the best way to select the subdomains, but also the most costly as it requires knowing all the subdomains' performances.
    \item \emph{Combinatorial Testing (CT)} \cite{nie2011survey} aims to test a limited number of combinations that cover well the search space. In particular, we use n-wise testing from the library allpairspy \cite{allpairspy}.  We vary n from 2 (pairwise testing) to 5 (because we have 5 attributes). This approach was used by \cite{metzen2023identification}.
\end{itemize}

\paragraph{Methods details}
We test two different approaches:
\begin{itemize}
    \item \emph{Genetic algorithm (GA)} We use a population size of 20 and the library pymoo \cite{pymoo}.
    \item \emph{Bayesian optimization (BO)} The predictor takes a one-hot embedding of the subdomain attributes as input and predicts the accuracy. We tested Random Forest Regressors (RFR) \cite{breiman2001random}, Linear Regression (LR),  Lasso \cite{lasso}, and Support Vector Regression (SVR) \cite{svr} using scikit-learn \cite{sklearn_api}. We start with a pre-training on 10 random subdomains to lower the variability of each run.
\end{itemize}

\paragraph{Metrics}
We study the evolution of selected subdomain accuracies, average accuracy of selected subdomains, and coverage of the 10\% lowest accuracies subdomains. We also show a histogram of the subdomain accuracies for a fixed number of explored subdomains. We use the Spearman rank correlation to evaluate the quality of the predictors.

\subsection{Evaluating all subdomains for reference}
\label{subsec:eval_all}

To validate our approach, we evaluate the performance of all subdomains and save the results as shown in Table \ref{tab:eval_data}. Because all evaluation results are pre-computed, benchmarking the different selection functions is done by replacing the generation and evaluation parts with a simple table look-up. This allows us to compare different selection functions quickly. This validation ensures that subsequent work can use our findings to reduce the number of evaluations. We generated 50 valid images for each of the 1032 subdomains. It took approximately 200 hours to generate all images on one NVIDIA V100 GPU. Sometimes, hundreds of images had to be generated to obtain 50 valid ones after filtering. The expected evaluation time of one subdomain is 12 minutes, or 1 hour for 5 subdomains. The results below show the number of subdomains explored as the x-axis. Still, we could have used an estimated computing time by using the value of around 12 minutes per subdomain evaluated.

Figure \ref{fig:samples} shows samples of generative images with their input prompt. While not perfect depictions of dogs, they are close enough to benchmark the classifier. Some images clearly depict dogs, yet the classifier fails to identify them. This highlights some of its limits.

\subsection{Benchmarking the selection functions}
\label{subsec:benchmark_selection}

The main goal of selection functions is to identify subdomains with low accuracy quickly. To measure this, we show the evolution of different metrics during the exploration in Figure \ref{fig:acc_plots}. The main conclusion is that combinatorial testing (n-wise testing with $n \in \{2, 3, 4, 5\}$) is not much better than random selection. Also, it has the disadvantage of restricting the number of subdomains selected: we cannot tune this number.  GA is much better, and BO is even better. BO can successfully identify all the 10\% most critical subdomains (with lowest accuracies) after evaluating $\approx$ 40\% of all subdomains. This also proves that subdomain performance can be precisely inferred from the domain attributes. This means that classifier failures can be explained from the attributes, providing interesting insights into the classifier decision process.

Figure \ref{fig:hist_acc} details a specific step in the evaluation process when the number of subdomains is equal to 61 (which is the number of subdomains selected by 3-wise testing). This also shows a clear advantage for GA and BO in quickly identifying low-accuracy subdomains.

Figure \ref{fig:spearman} shows that the four predictors perform similarly well. We choose Lasso as a predictor for BO because it is the best method for small training set sizes. Indeed, the beginning of the exploration, when the data is limited, is particularly important. Furthermore, it showed less variability than, for example, random forests.

\subsection{Qualitative analysis of classifier failures}
\label{subsec:quali}

\begin{figure}[hb]
    \centering
    \scalebox{0.75}{
\begin{tikzpicture}

\definecolor{darkslategray38}{RGB}{38,38,38}
\definecolor{lightgray204}{RGB}{204,204,204}

\begin{axis}[
axis line style={darkslategray38},
colorbar,
colorbar style={ylabel={}},
colormap={mymap}{[1pt]
  rgb(0pt)=(0.83921568627451,0.152941176470588,0.156862745098039);
  rgb(1pt)=(0.951036373240266,0.941442264351832,0.94133244907644)
},
height=8cm,
point meta max=0.906666656335195,
point meta min=0.489230751991272,
tick align=outside,
tick pos=left,
width=8cm,
x grid style={lightgray204},
xlabel=\textcolor{darkslategray38}{location},
xmin=0, xmax=7,
xtick style={color=darkslategray38},
xtick={0.5,1.5,2.5,3.5,4.5,5.5,6.5},
xticklabel style={rotate=80.0},
xticklabels={
  at the beach,
  in a garden,
  in the city,
  in the desert,
  in the forest,
  in the mountains,
  inside a house
},
y dir=reverse,
y grid style={lightgray204},
ylabel=\textcolor{darkslategray38}{time / weather},
ymin=0, ymax=7,
ytick style={color=darkslategray38},
ytick={0.5,1.5,2.5,3.5,4.5,5.5,6.5},
yticklabels={
  day / foggy,
  day / raining,
  day / snowing,
  day / sunny,
  night / foggy,
  night / raining,
  night / snowing
}
]
\addplot graphics [includegraphics cmd=\pgfimage,xmin=0, xmax=7, ymin=7, ymax=0] {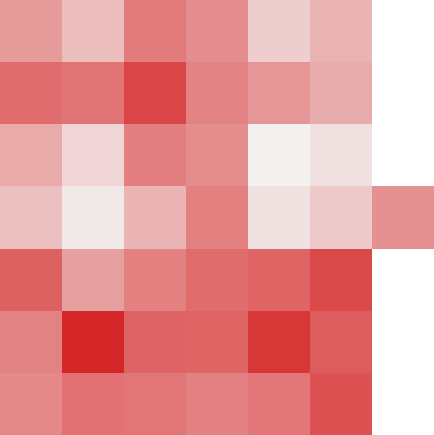};
\end{axis}

\end{tikzpicture}
        }
    \caption{Heatplot displaying the average accuracies for different combinations of weather and location.}
    \label{fig:heatplot}
\end{figure}

The main focus of our work is to efficiently detect the attributes with the most impact on classification performance. However, this subsection suggests what kind of qualitative assessment it allows. We use the BO approach and allow the exploration of 300 subdomains. Figure \ref{fig:barplot} shows the average accuracies for each attribute's value. This shows the impact of each attribute individually but does not show the impact of combinations of attributes. Figure \ref{fig:heatplot} displays the impact of all the possible combinations of the attributes of weather and location.

\section{Limitations}

\begin{figure}[ht]
    \centering
    \begin{subfigure}{.15\textwidth}
        \centering
        \includegraphics[width=\linewidth]{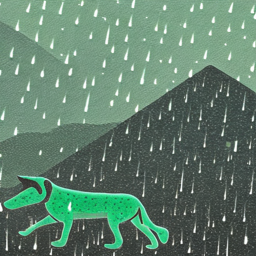}
    \end{subfigure}
    \hfill
    \begin{subfigure}{.15\textwidth}
        \centering
        \includegraphics[width=\linewidth]{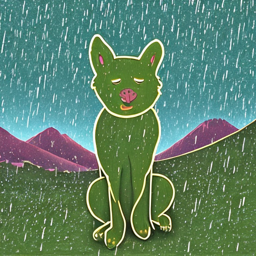}
    \end{subfigure}
    \hfill
    \begin{subfigure}{.15\textwidth}
        \centering
        \includegraphics[width=\linewidth]{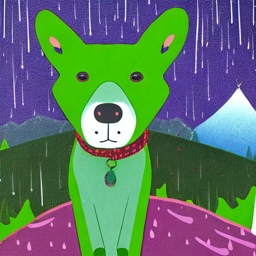}
    \end{subfigure}
    \vfill
    \vspace{3mm}
    \begin{subfigure}{.15\textwidth}
        \centering
        \includegraphics[width=\linewidth]{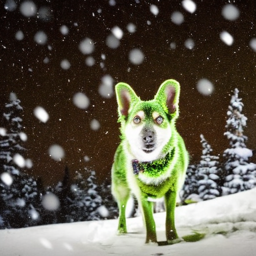}
    \end{subfigure}
    \hfill
    \begin{subfigure}{.15\textwidth}
        \centering
        \includegraphics[width=\linewidth]{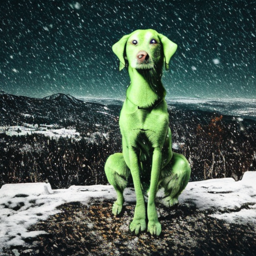}
    \end{subfigure}
    \hfill
    \begin{subfigure}{.15\textwidth}
        \centering
        \includegraphics[width=\linewidth]{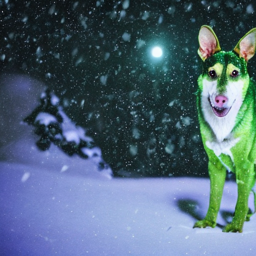}
    \end{subfigure}
    \caption{In the top row, images are generated with the prompt "A front view of a green dog in the mountains, during the night, it is raining.". They are mostly in a cartoon style. In the bottom row, the same prompt, but "raining" has been replaced by "snowing". The phenomenon disappears. Is this a generator failure? Careful prompt engineering, e.g., adding "a realistic image", is required to ensure alignment between the textual prompt, images, and what we expect.}
    \label{fig:samples_bias}
\end{figure}

Benchmarking classifiers with generative models has limitations as observed by other work \cite{wiles2022discovering,metzen2023identification}. There can be occasional misalignments between the prompt and the image due to bias or language limitations. For instance, in this work, we observed that the viewpoint attribute is sometimes not the one requested. We also observed generator failures for a few specific subdomains, e.g., nearly all images for "A front view of a green dog in the mountains, during the night, it is raining." are in a cartoon style, which is not the case for snowing, see Figure \ref{fig:samples_bias}. Prompt engineering is required to allow a rigorous benchmark of the classifier.  Also, generated images do not cover everything possible in the real world. Our approach tackles the computing time problem. Its main limitation is that there is no guarantee that a good selection function will identify \emph{all} problematic subdomains for an incomplete exploration. For instance, a subdomain might be difficult for completely different reasons than the others. Thus, a selection based on learning a relation between subdomain attributes and performance might miss it.

\section{Conclusion and perspectives}

Text-to-Image models have great potential to be a useful tool for benchmarking image classifiers by generating images of failure cases. However, since the highest quality generators are based on diffusion models, their high inference time prevents large-scale image synthesis for advanced evaluation. This work starts from an evaluation domain described by textual attributes. To efficiently explore the critical attribute combinations that cause classifier failures, we propose to create an iterative process that alternates image generation, classifier evaluation, and attribute selection. We compare different selection functions and show that all of them outperform the method used in a previous work.

We believe that our work can be further improved by using NAS methods, taking advantage of low-fidelity evaluations. For example, in our case, the accuracy could be estimated with 20 images. The method would then use these low-fidelity evaluations to decide which combination is worth testing with high-fidelity, say 200 images. In addition, for more complex problems, one can use word embeddings such as language models instead of one-hot embeddings of finite attributes.
Our work can potentially improve the benchmarking of image classifiers with text-to-image models, as it addresses a major limitation: computational time. It allows the exploration of larger domains and more precise estimates of accuracies, class probabilities, and failures.

\section*{Acknowledgments}
This work has been supported by the French government under the "France 2030” program, as part of the SystemX Technological Research Institute.\\
This work was granted access to the HPC/AI resources of IDRIS under the allocation 2024-AD011013372R2 made by GENCI.

\clearpage
{
    \small
    \bibliographystyle{ieeenat_fullname}
    \bibliography{biblio}
}


\end{document}